\documentclass[10pt,twocolumn,letterpaper]{article}
\usepackage{cvpr}
\cvprfinalcopy

\usepackage{epsfig}
\usepackage{graphicx}
\usepackage{amsmath}
\usepackage{mathtools}
\usepackage{amssymb}
\usepackage{tabularx}
\usepackage{inputenc}
\usepackage{float}
\usepackage{subfigure}
\usepackage{color}
\usepackage{algorithmic}
\usepackage{algorithm}
\usepackage{setspace}
\usepackage[pagebackref=true,breaklinks=true,colorlinks,bookmarks=false]{hyperref}
\ifcvprfinal\hypersetup{
    pdftitle={Why do linear SVMs trained on HOG features perform so well?},
    pdfauthor={Hilton Bristow and Simon Lucey}, 
    pdfsubject={CVPR 2014},
    pdfcreator={LaTeX}, 
    pdfproducer={IEEE},
    pdfkeywords={big data} {second-order} {features} {HOG} {CVPR} {2014} {paper},
}\fi

\newcommand{\x}{\mathbf{x}}

\newcommand{\e}{\mathbf{e}}
\newcommand{\g}{\mathbf{g}}

\newcommand{\w}{\mathbf{w}}

\renewcommand{\b}{\mathbf{b}}

\newcommand{\R}{\mathbb{R}}
\newcommand{\M}{\mathbf{M}}
\newcommand{\B}{\mathbf{B}}
\newcommand{\G}{\mathbf{G}}

\renewcommand{\L}{\mathbf{L}}
\renewcommand{\P}{\mathbf{P}}
\renewcommand{\S}{\mathbf{S}}
\renewcommand{\H}{\mathbf{G}}

\newcommand{\I}{\mathbf{I}}

\newcommand{\D}{\mathbf{D}}

\renewcommand{\vec}{\operatorname{vec}}
\newcommand{\kron}{\otimes}
\renewcommand{\dot}{\odot}
\newcommand{\conv}{\ast}

\newcommand{\eqn}[1]{Equation (\ref{#1})}
\newcommand{\fig}[1]{Figure \ref{#1}}

\newenvironment{packed_item}{
\begin{itemize}
  \setlength{\itemsep}{1pt}
  \setlength{\parskip}{0pt}
  \setlength{\parsep}{0pt}
}{\end{itemize}}

\renewcommand{\subsection}[1]{\vspace{5pt} \noindent \textbf{#1:}}



\usepackage{xspace}
\makeatletter
\DeclareRobustCommand\onedot{\futurelet\@let@token\@onedot}
\def\@onedot{\ifx\@let@token.\else.\null\fi\xspace}

\def\ie{\emph{i.e}\onedot} 
 
\def\etc{\emph{etc}\onedot} 
 
\def\etal{\emph{et al}\onedot}
\makeatother
\newtheorem{remark}{Remark}
\newtheorem{hypothesis}{Hypothesis}

\title{Why do linear SVMs trained on HOG features perform so well?}
\author{\href{mailto:hilton.bristow@gmail.com}{Hilton Bristow}\textsuperscript{1} and 
             \href{mailto:slucey@cs.cmu.edu}{Simon Lucey}\textsuperscript{2}\\
\textsuperscript{1}Queensland University of Technology, Australia\\
\textsuperscript{2}Carnegie Mellon University, USA}

\def\cvprPaperID{30}

\begin{document}
\maketitle


\begin{abstract}
Linear Support Vector Machines trained on HOG features are now a de facto standard across many visual perception tasks. Their popularisation can largely be attributed to the step-change in performance they brought to pedestrian detection, and their subsequent successes in deformable parts models. This paper explores the interactions that make the HOG-SVM symbiosis perform so well. By connecting the feature extraction and learning processes rather than treating them as disparate plugins, we show that HOG features can be viewed as doing two things: (i) inducing capacity in, and (ii) adding prior to a linear SVM trained on pixels. From this perspective, preserving second-order statistics and locality of interactions are key to good performance. We demonstrate surprising accuracy on expression recognition and pedestrian detection tasks, by assuming only the importance of preserving such local second-order interactions.
\end{abstract}


\section{Introduction}
Despite visual object detectors improving by leaps and bounds in the last decade, they still largely rely on the same underlying principles: linear Support Vector Machines (SVMs) trained on Histogram of Oriented Gradient (HOG) features. When HOG features were introduced, they improved upon existing methods for pedestrian detection by an order of magnitude~\cite{dalal_triggs_CVPR_2005}. Since then, the HOG-SVM pipeline has been used in the deformable parts model of~\cite{felzenszwalb_PAMI_2010}, its derivatives, and high achievers of the PASCAL VOC challenge. A keyword search for ``HOG'' and ``SVM'' in the PASCAL 2012 results page returns 25 and 18 hits respectively. HOG is now complicit in detection pipelines across almost every visual detection/classification task~\cite{dalal_triggs_CVPR_2005,deniz_PR_2011,xiao_CVPR_2010,yang_CVPR_2011}. 

In this paper we peel back the layers of complexity from HOG features to understand what underlying interactions drive good performance. In particular,
\begin{packed_item}
\item HOG features can be viewed as an affine weighting on the margin of a quadratic kernel SVM,
\item underlying this prior and added capacity is the preservation of local pixel interactions and second-order statistics,
\item using these foundational components alone, we show it is possible to learn a high performing classifier, with no further assumptions on images, edges or filters.
\end{packed_item}


\section{Representing HOG features as a linear transform of pixels}
\begin{figure*}
\includegraphics[width=\textwidth,trim=0 0 0 0]{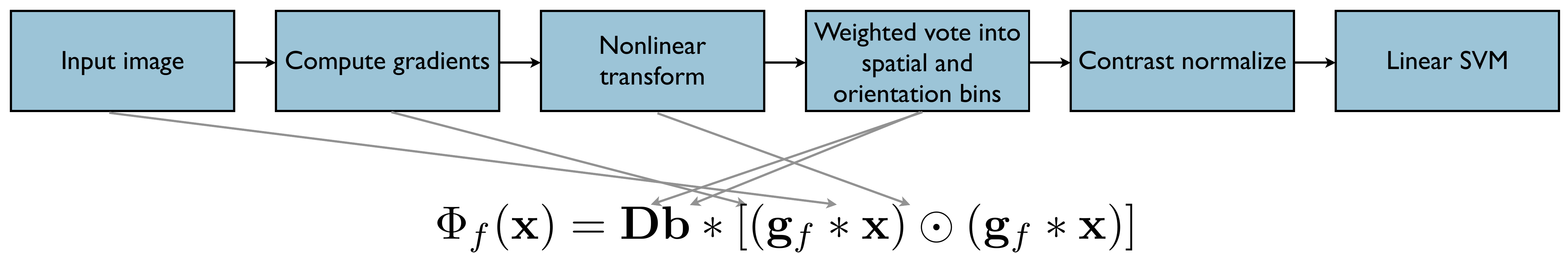}
\caption{An illustration of the HOG feature extraction process and how each component maps to our reformulation. Gradient computation is achieved through convolution with a bank of oriented edge filters. The nonlinear transform is the pointwise squaring of the gradient responses which removes sensitivity to edge contrast and increases edge bandwidth. Histogramming can be expressed as blurring with a box filter followed by downsampling.}
\label{fig:pipeline}
\end{figure*}

HOG features can be described as taking a nonlinear function of the edge orientations in an image and pooling them into small spatial regions to remove sensitivity to exact localisation of the edges. A pictorial representation of this pipeline is shown in \fig{fig:pipeline}. This type of representation has proven particularly successful at being tolerant to non-rigid changes in object geometry whilst maintaining high selectivity~\cite{dicarlo_NEURON_2012}. 

The exact choice of non-linear function is discretionary, however Dalal and Triggs try $f(x) = |x|$ and $f(x) = |x|^2$. Both functions remove the edge direction, leaving only a function of the magnitude. In this paper, we choose to use the square of edge directions. Our choice is motivated by a number of reasons. First, the square function leads to greater flexibility in manipulation, which we show becomes vital in our later reformulation. Second, there are proponents of the squaring function (referred to as ``square'' or ``L2'' pooling) in convolutional network literature~\cite{bergstra_technical_2009,kavukcuoglu_CVPR_2009}, whose architectures have a similar pipeline to HOG features per layer, and have shown exciting results on large-scale visual recognition challenges~\cite{krizhevsky_NIPS_2012}. Finally, the square function has a good basis in statistical models of the primary visual cortex~\cite{hyvarinen_network_2007}.

Following the notation of~\cite{bristow_ECCV_2012}, the transform from pixels to output in HOG features can be written as,
\begin{align}
\Phi_f(\x) = \D \b \conv \left[ (\g_f \conv \x) \dot (\g_f \conv \x) \right] \;,
\end{align}
where a vectorized input image $\x \in \R^D$ is convolved with an oriented edge filter $\g_f$~\cite{Daugman_OSA_1985}, rectified through the Hadamard operator (pointwise square), then finally blurred with $\b$ and downsampled by the sparse selection matrix $\D$ to achieve pooling/histogramming. Performing this operation over a bank of oriented edge filters and concatenating the responses leads to the final descriptor,
\begin{align}
\Phi(\x) = \left[ \Phi_1(\x) \;\; \Phi_2(\x) \;\; \dots \;\; \Phi_F(\x) \right] \;. \label{eqn:squarepool}
\end{align}

This reformulation omits the contrast normalization step, which we address later in the piece. We showed previously in~\cite{bristow_ECCV_2012} that each sub-descriptor can be expressed in the form,
\begin{align}
\Phi_f(\x) = \D \B \M (\G_f \kron \G_f)(\x \kron \x) \;,
\end{align}
where $\M$ is a selection matrix and capitalized $\B$, $\G$ are matrix forms of their convolutional prototypes. The full response to a bank of filters can be written as,
\begin{align}
\Phi(\x) = \L(\x \kron \x) \;, \label{eqn:Lxx}
\end{align}
where the projection matrix $\L$ is formed by concatenating the bank,
\begin{align}
\L = \begin{bmatrix}
\B\M (\G_{1} \kron \H_{1})\\
\vdots \\
\B\M (\G_{F} \kron \H_{F}) \\
\end{bmatrix} \;.
\end{align}

Under this reformulation, HOG features can be viewed as an affine weighting of quadratic interactions between pixels in the image. This suggests that the good performance of HOG features is down to two things: (i) second-order interactions between pixels, and (ii) the choice of prior that forms the affine weighting $\L$.

\begin{figure*}
\center
\includegraphics[width=\textwidth]{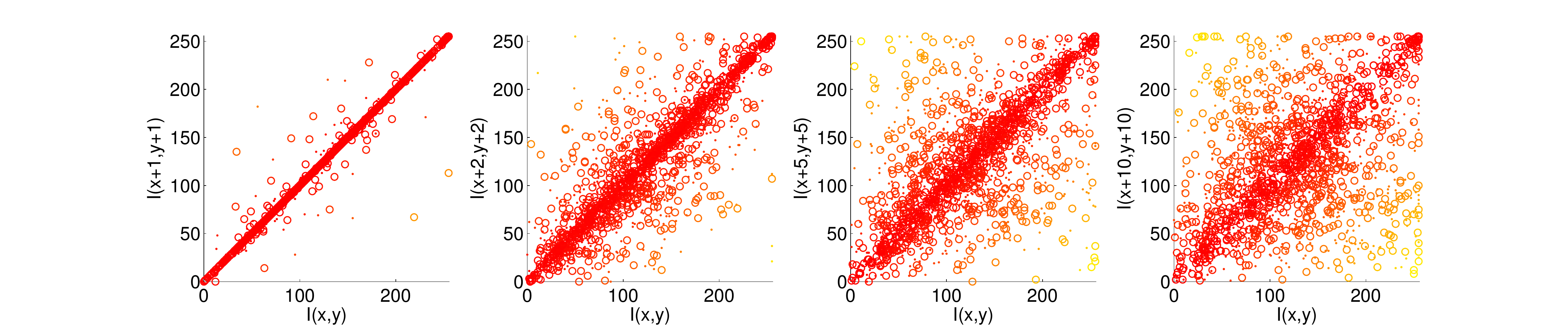}
\caption{An illustration of the locality of pixel correlations in natural images. Whilst a single pixel displacment exhibits strong correlations in intensity, there are few discernible correlations beyond a 5 pixel displacement. Locality is also observed in the human visual system, where cortical cells have finite spatial receptive fields.
\label{fig:pixelcorrelation}}
\end{figure*}


\section{Capacity of the Classifier}
When used in a support vector classification setting, the projection matrix $\L$ can be absorbed into the \mbox{margin},
\begin{alignat}{4}
\w^{*} =&\;\arg\min_{\mathclap{\w,\xi_{i} \geq 0}} && \frac{1}{2} \w^{T}(\L^{T} \L)^{-1}\w + C \sum_{i=1}^{l} \xi_{i} \label{eqn:SVM-primal-new} \\
           &\;\; \mbox{subject to} \quad &&  y_{i} \w^{T} (\x_{i} \kron \x_{i}) \geqslant 1 - \xi_{i}, \quad  i = 1 \ldots l \nonumber
\end{alignat}
leaving as the data term only interactions between the weights $\w$ and the Kronecker expansion of the image. 

\begin{remark}
When the weighting is the identity matrix, (\ie $\L = \I$), \eqn{eqn:SVM-primal-new} reduces to an SVM with a unary quadratic kernel.
\end{remark}

The first part of the HOG story, the induced prior, can thus be quantified as being quadratic. The weighting matrix is therefore intrinsically a prior on the margin of a quadratic kernel SVM.


\section{Secord-Order Interactions}
The term $(\x \kron \x)$ in \eqn{eqn:Lxx} can alternatively be written as,
\begin{align}
(\x \kron \x) = \vec(\x \x^T) \;,
\end{align}
which is the vectorized covariance matrix of all pixel interactions. \cite{tuzel_ECCV_2006} showed that the covariance of a local image distribution is often enough to discriminate it from other distributions. However, when dealing with high-dimensional distributions, computing a full-rank covariance matrix is often difficult. \cite{hariharan_ECCV_2012} circumvent this problem by assuming stationarity of background image statistics (a translated image is still an image), as well as limiting the bandwidth of interactions between pixels. Simoncelli~\cite{simoncelli_NEUROSCIENCE_2001} showed that these assumptions are reasonable, since correlations between pixels fall quickly with distance (see \fig{fig:pixelcorrelation}).

To improve conditioning and prevent overfitting the classifier to hallucinated interactions, we consider the most general set of \emph{local} second-order features: the set of all local unary second-order interactions in an image,
\begin{align}
\Psi(\x) = [\vec \{ \S_{1}(\x) \}^{T},  \ldots, \vec \{ \S_{D}(\x)
\}^{T} ]^{T}
\label{eqn:lcf} 
\end{align}
where,
\begin{align}
\S_{i}(\x) = \P_{i} \x \x^{T} \P_{i}^{T} \;,
\label{eqn:lcfsub}
\end{align}
$\P_{i}$ is simply an~$M \times D$ matrix that extracts an~$M$ pixel local region centred around the~$i$th pixel of the image~$\x$. By retaining local second-order interactions, the feature length grows from $D$ for raw pixels to~$M^{2} D$.

Fortunately, inspection of~\eqn{eqn:lcf} reveals a large amount of redundant information. This redundancy stems from the re-use of pixel interactions in surrounding local pixel locations. Taking this into account, and without loss of information, one can compact the local second-order feature to $MD$ elements, so that~\eqn{eqn:lcf} becomes,
\begin{align}
\Psi^{*}(\x) = 
\begin{bmatrix} (\e_{1} * \x)^{T} \circ \x^{T} \\
\vdots \\
(\e_{M} * \x)^{T} \circ \x^{T} \end{bmatrix} \;. 
\label{eqn:compact}
\end{align}
where~$\{\e_{m}\}_{m=1}^{M}$ is the set of~$M$ impulse filters that encode the local interactions in the signal. 

\section{Local Second-Order Interactions}
\begin{figure*}
\center
\includegraphics[width=\textwidth]{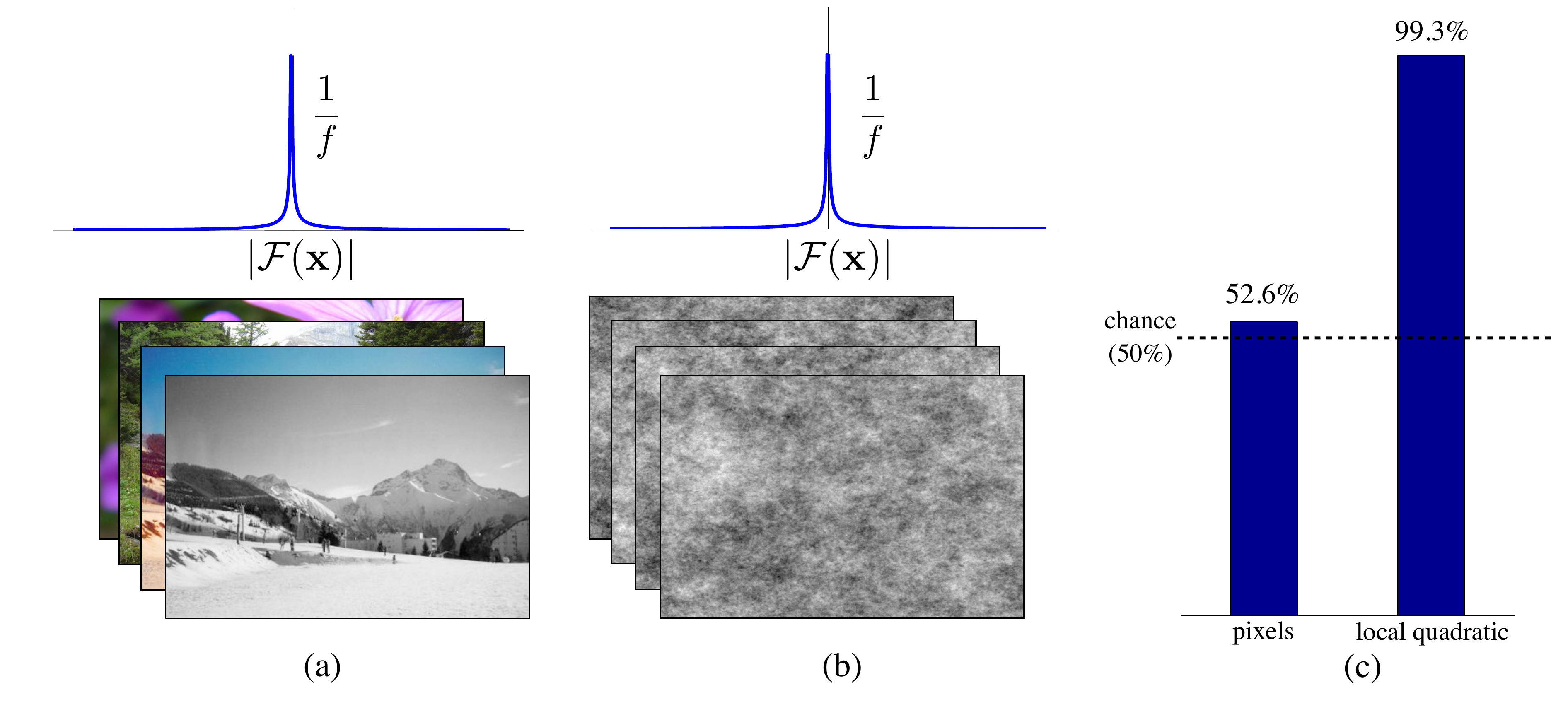}
\caption{Thought-experiment setup. (a) contains an ensemble of samples drawn from the space of natural images with a \mbox{$\frac{1}{f}$ frequency} spectrum, (b) contains an ensemble of samples drawn from a random noise distribution \emph{with the same $\frac{1}{f}$ frequency spectrum}. (c) We train two linear classifiers to distinguish between ``natural'' or ``noise.'' The pixel-based classifier does not have the capacity to discriminate between the distributions. The classifier which preserves local quadratic pixel interactions almost perfectly separates the two distributions. 
\label{fig:noise_natural}}
\end{figure*}

Consider two classes A and B. Class A represents the distribution of all natural images. Class B represents a noise distribution which has the same frequency spectrum as natural images, namely $\frac{1}{f}$~\cite{simoncelli_NEUROSCIENCE_2001}. Both distributions are power normalized. We sample 25000 training and testing examples from each class, and train two classifiers: one preserving the raw pixel information and one preserving \emph{local second-order interactions} of the pixels. The goal of the classifiers is to predict ``natural'' or ``noise.'' An illustration of the experimental setup and the results are presented in \fig{fig:noise_natural}. The pixel classifier fails to discriminate between the two distributions. There is no information in either the spatial or Fourier domain to linearly separate the classes (\ie the distributions overlap). By preserving local quadratic interactions of the pixels, however, the classifier can discriminate natural from synthetic almost perfectly. 

Whilst the natural image and noise distributions have the same frequency spectra, natural images are not random: they contain structure such as lines, edges and contours. This experiment shows that image structure is inherently local, and more importantly, that local second-order interactions of pixels can exploit this structure. Without encoding an explicit prior on edges, pooling, histogramming or blurring, local quadratic interactions have sufficient capacity to exploit the statistics of natural images, and separate them from noise. 

\section{Replacing prior with posterior: \\learning over pixels}
Quadratic kernel SVMs have not historically performed well on recognition tasks when learned using pixel information. The image prior that HOG encodes, and the affine weighting that it can be distilled into, is integral to obtaining good generalisation performance. We know, however, that a prior is simply used to reflect our belief in the posterior distribution in the absence of actual data. In the case of HOG, the prior encodes insensitivity to local non-rigid deformations so that the entire space of deformation does not need to be sampled to make informed decisions.

This is usually a reasonable assumption to make, since sampling the posterior sufficiently may be infeasible. Take, for example, the task of pedestrian detection. The full posterior comprises all possible combinations of pose, clothing, lighting, race, gender, identity, background and any other attribute that manifests in a change to the visual appearance of a person. Multi-scale sliding window HOG detectors (with pose elasticity in the case of DPMs) work to project out as much of this intra-class variation as possible.

Is it possible to learn a detector using only the assumptions that underlie HOG features: the preservation of local second-order interactions? How much data is required to render the HOG prior unnecessary, and what sort of data is required? Can the data just be perturbations of the training data? Does the resulting classifier learn anything more specialized than one learned on HOG features? In the following section we aim to provide answers to these questions.

\begin{hypothesis}
If HOG features are doing what we intend - providing tolerance to local geometric misalignment - then it should be possible to reproduce their effects by sampling from a sufficiently large dataset containing geometrically perturbed instances of the original training set.
\end{hypothesis} 

\subsection{Learned Features}
It is the firm belief of many that learned features are the way forward~\cite{hinton_CS_2007}. Convolutional network literature has heavily relied on learned features for a number of years already~\cite{kavukcuoglu_NIPS_2010,lee_ICML_2009}. We take feature learning to its most primitive form, setting only the capacity and distribution of features, and letting the classifier learn the rest.


\section{Methods}
We designed an experiment where we could control the amount of geometric misalignment observed between the training and testing examples. We used the Cohn Kanade+ expression recognition dataset, consisting of 68-point landmark, broad expression and FACS labels across 123 subjects and 593 sequences. Each sequence varies in length and captures the neutral and peak formation of facial expression. In this paper we consider only the task of broad expression classification (\ie we discard FACS encodings). To test the invariance of different types of features to geometric misalignment, we first register each training example to a canonical pose, then synthesize similarity warps of the examples with increasing RMS point error.

\subsection{Why Faces?}
HOG features have been used across a broad range of visual recognition tasks, including object recognition, scene recognition, pose estimation, \etc. Faces are unique, however, since they are a heavily studied domain with many datasets containing subjects photographed under controlled lighting and pose conditions, and labelled with ground-truth facial landmarks. This enables a great degree of flexibility in experimental design, since we can programatically set the amount of geometric misalignment observed while controlling for pose, lighting, expression and identity.

\begin{figure}
\center
\includegraphics[width=\columnwidth,trim=0 10 0 0]{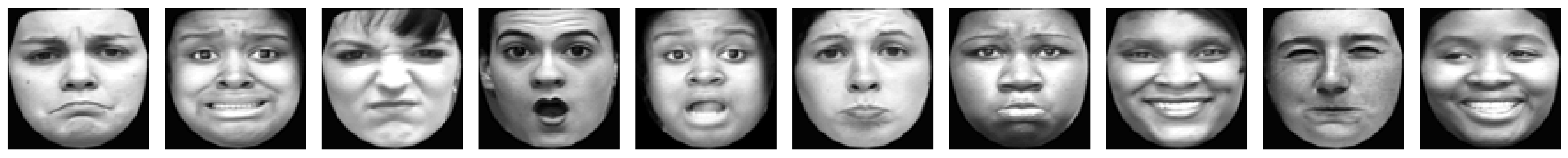}
\\(a) 0-pixel error\\
\includegraphics[width=\columnwidth,trim=0 10 0 -40]{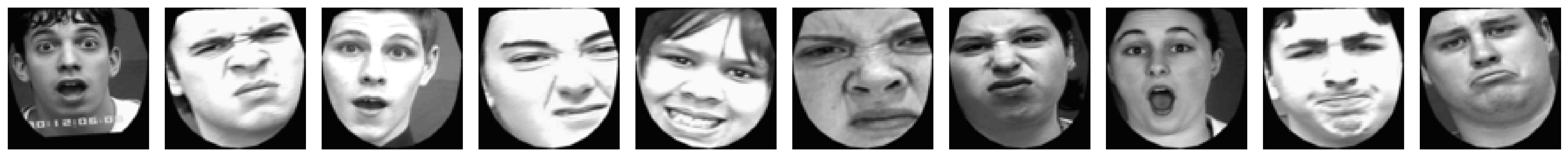}
\\(b) 10-pixel error\\
\caption{Illustrative examples of subjects from the Cohn Kanade+ dataset with (a) zero registration error, and (b) 10 pixels of registration error.
\label{fig:exemplars}}
\end{figure}

We synthesize sets with 300, 1500, 15000 and 150000 training examples. The larger the synthesized set, the greater the coverage of geometric variation. We use HOG features according to Felzenszwalb \etal~\cite{felzenszwalb_PAMI_2010} with 18 orientations and a spatial aggregation size of 4. For the reformulation of \eqn{eqn:squarepool}, we use Gabor filters with 18 orientations at 4 scales, and a $4 \times 4$ blur kernel. The local quadratic features have a spatial support equal to the amount of RMS point error (\ie at 10 pixels error, correlations are collected over $10 \times 10$ regions). All training images are $80 \times 80$ pixels and cropped around only the faces. \fig{fig:exemplars} illustrates the degree of geometric misalignment introduced.

\subsection{Contrast Normalization}
So far we have neglected to mention contrast normalization, the final stage in the HOG feature extraction pipeline, and a component that has traditionally received much attention, particularly in neuroscience literature~\cite{carandini_NATURE_2012}. Whilst contrast normalization is an integral part of HOG features, we do not believe it plays a significant role in their invariance to geometry. Since our model does not explain the highly nonlinear process of contrast normalization, we instead power normalize images in the expression recognition experiment, and pre-whiten images in the pedestrian detection experiment.

\subsection{Learning}
The storage requirements of local quadratic features quickly explode with increasing geometric error and synthesized examples. At 10 pixels RMS error, 150000 training examples using local quadratic features takes 715 GB of storage. To train on such a large amount of data, we implemented a parallel support vector machine~\cite{boyd_book_2010} with a dual coordinate descent method as the main solver~\cite{hsieh_ICML_2008}. Training on a Xeon server using 4 cores and 24 GB of RAM took between 1 -- 5 days, depending on problem size. We used multiple machines to grid search parameters and run different problem sizes.

\begin{figure*}
\center
\includegraphics[width=\textwidth,trim=0 0 0 20,clip]{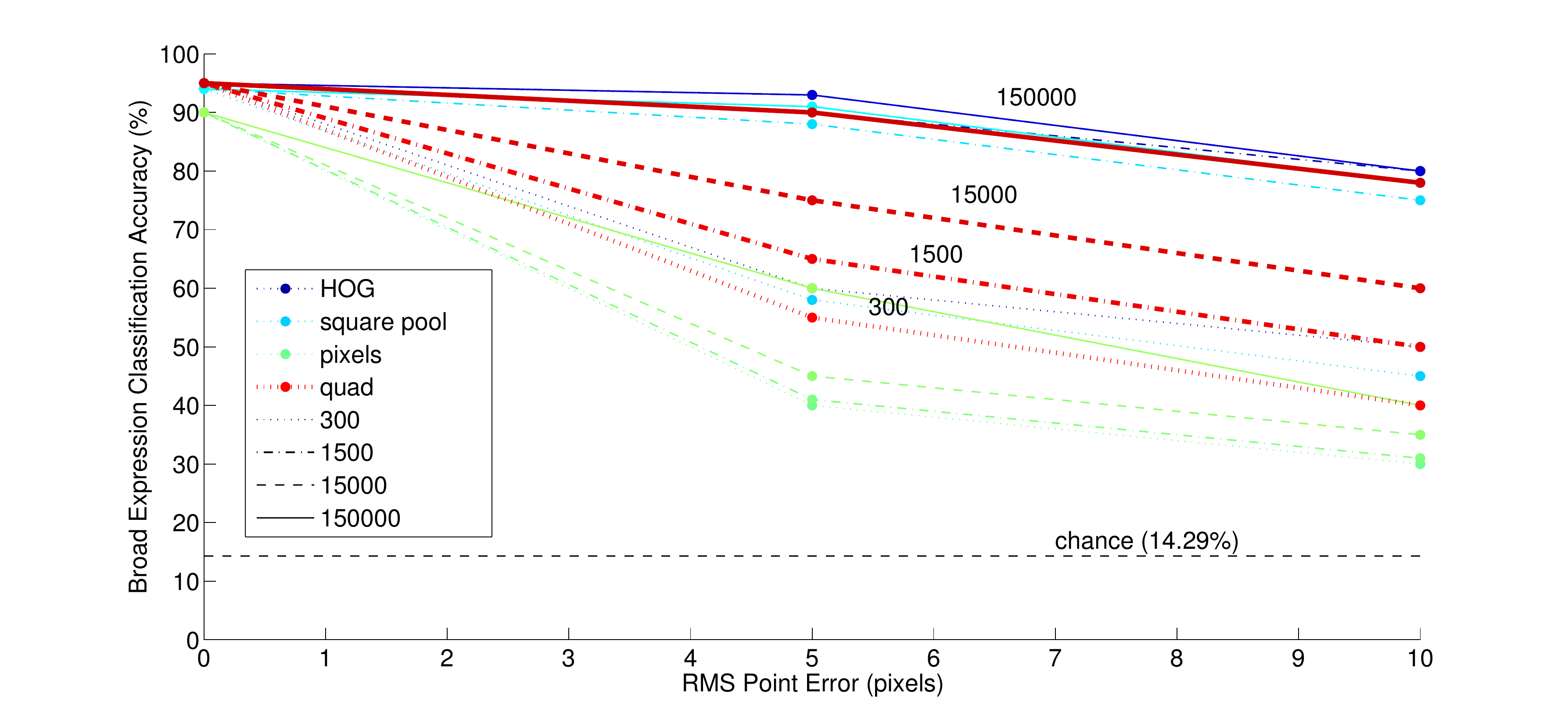}
\caption{Broad expression classification accuracy for different feature representations as a function of alignment error and amount of training data. For each feature representation we synthesized 300, 1500, 15000 and 150000 training examples. The held out examples used for testing \emph{always} come from an unseen identity. HOG features quickly saturate as the amount of training data increases. Quadratic features, shown in red, have poor performance with only a small number of synthesized examples, but converge towards the performance of HOG as the space of geometric variation is better spanned. Quadratic features appear to improve by roughly 10\% per order of magnitude of training data, until saturation.
\label{fig:main}}
\end{figure*}

\fig{fig:main} shows a breakdown of the results for synthesized sets of geometric variation. Pixels (shown in shades of green) perform consistently poorly, even with large amounts of data. HOG features (in blue, and reformulation in aqua) consistently perform well. The performance of HOG saturates after just 1500 training examples. \cite{zhu_BMVC_2012} talk about the saturation of HOG at length, noting that more data sometimes \emph{decreases} its performance. 

Local quadratic features (shown in red) have a marked improvement in performance with increasing amounts of data (roughly 10\% per order of magnitude of training data). Synthesizing variation used to be quite popular in some vision circles, particularly face recognition through the use of AAMs~\cite{zhang_PR_2009}, however it seems to have gone out of fashion in object recognition. Our results suggest that a model with sufficient capacity could benefit from synthesized data, even on general object recognition tasks (see Pedestrian Detection).

Only when the dataset contains $\ge 100000$ examples do the local quadratic features begin to model non-trivial correlations correctly. With 150000 training samples, local quadratic features perform within 3\% of HOG features. 

\subsection{Pedestrian Detection}
\label{sec:human}
We close with an example showing how the ideas of \emph{locality} and \emph{second-order} interactions can be used to learn a pedestrian detector. We don't intend to outperform HOG features. Instead we show that our insights are important for good performance, in contrast with a pixel-based classifier.

\begin{figure}
\center
\includegraphics[width=\columnwidth,trim=65 15 50 10,clip]{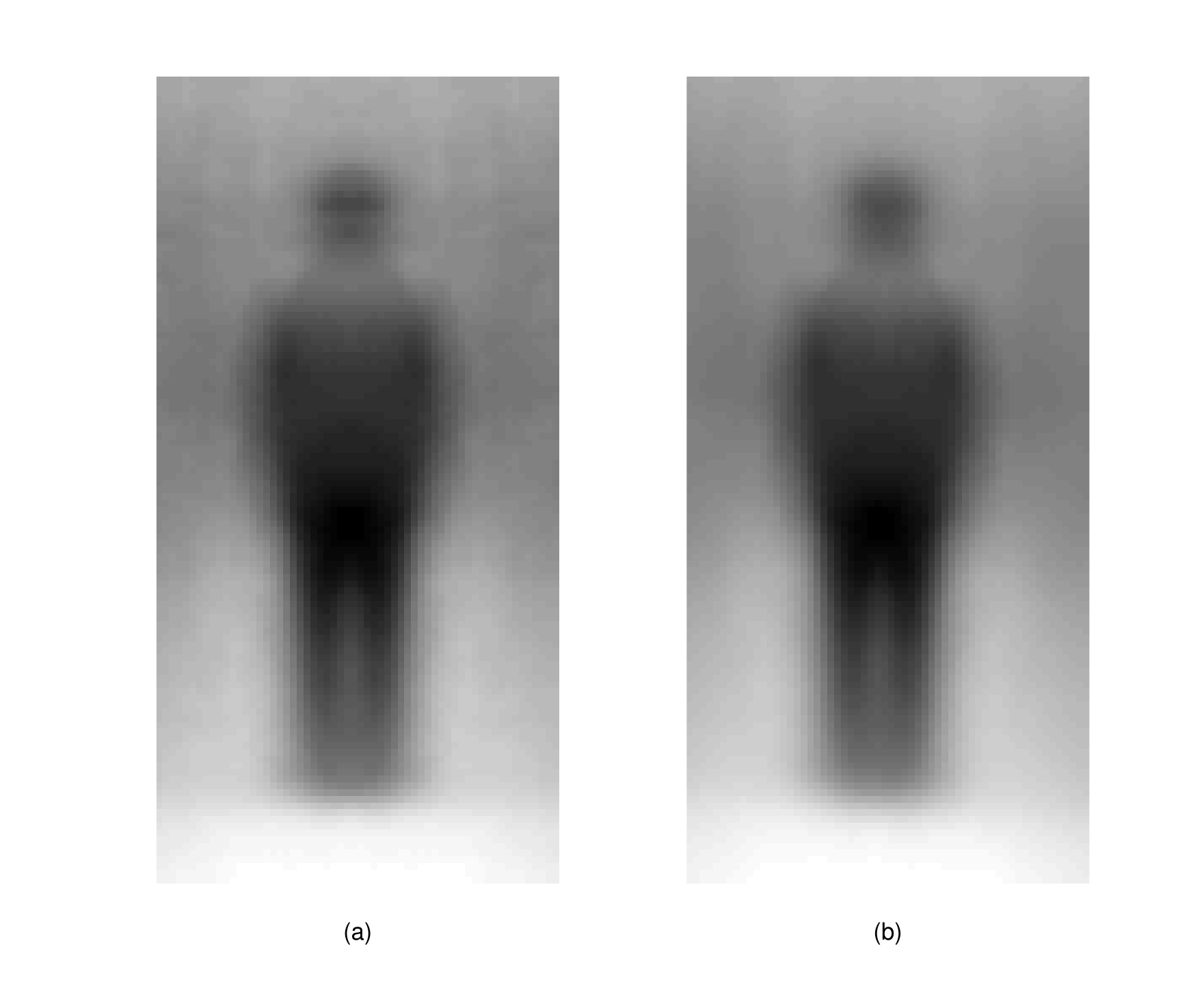}
\caption{The pixel mean of positive examples from the INRIA training set, (a) only, and (b) with 20 synthesized warps per example. The mean is virtually the same, suggesting that the synthesized examples are not adding ridig transforms that could be accounted for by a multi-scale sliding-window classifier. 
\label{fig:inria-mean-synth}}
\end{figure}

We follow a similar setup to our earlier expression recognition experiment on INRIA person. We generate synthetic similarity warps of each image, making sure they remain aligned with respect to translation. \fig{fig:inria-mean-synth} illustrates how the addition of synthesized examples does not change the dataset mean appreciably (misalignment would manifest as blur). We train the SVM on 40,000 positive examples and 80,000 negative examples, without hard negative mining.

\begin{figure}
\center
\includegraphics[width=\columnwidth,trim=10 0 10 0,clip]{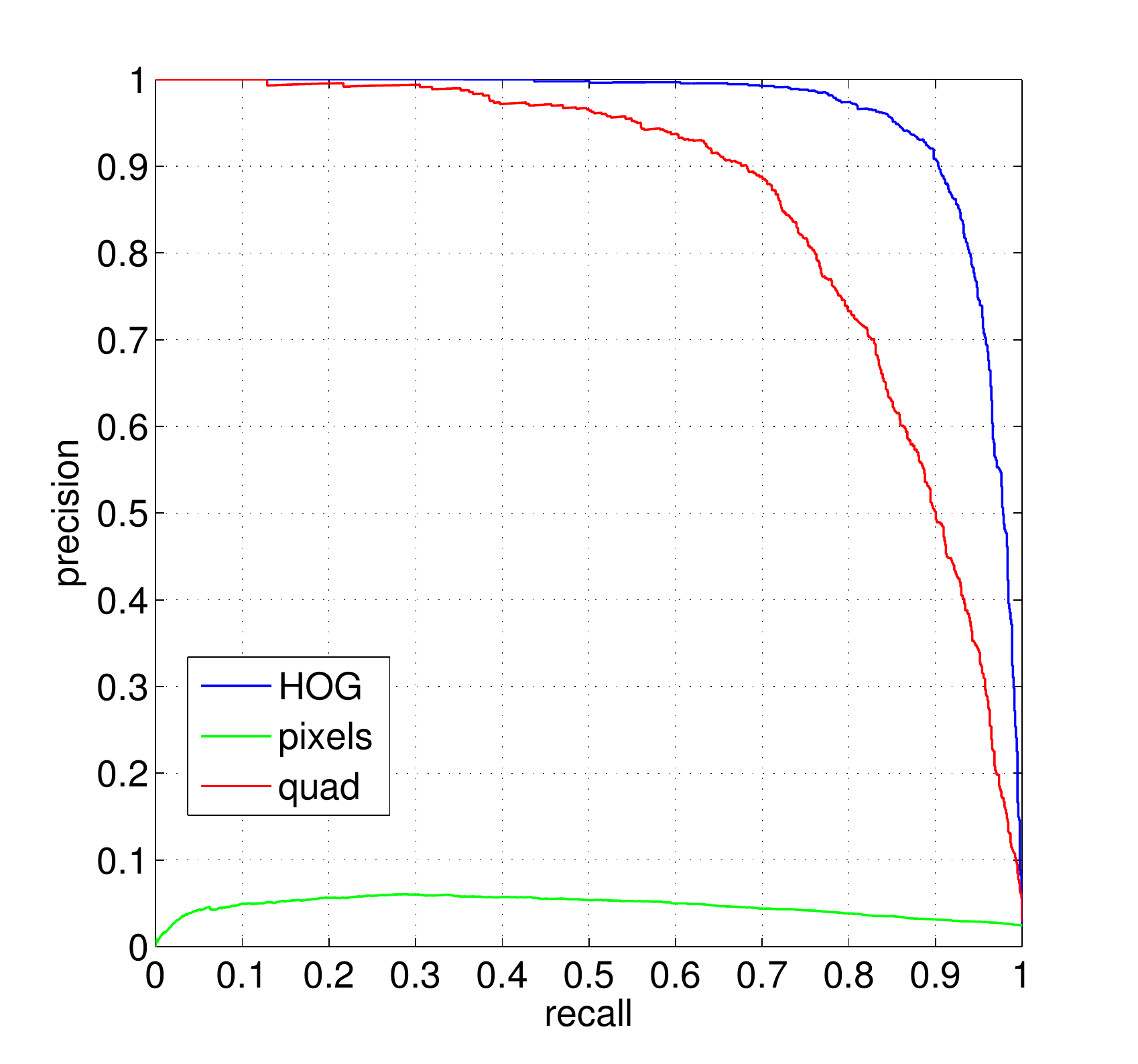}
\caption{Precision recall for different detectors on the \mbox{INRIA} person dataset. HOG performs well and pixels perform poorly, as expected. Local second-order interactions of the pixels (quad) perform surprisingly well, considering the lack of image prior and contrast normalization. The added capacity and locality constraints go a long way to achieving HOG-like performance. 
\label{fig:inria-results}}
\end{figure}

The results are striking. Unsurprisingly, the pixel-based classifier has high detection error, whilst the HOG classifier performs well. The local-quadratic classifier falls between the two, with an equal error rate of 22\%. The improved performance can be attributed solely to the added classifier capacity and its ties to correlations within natural image statistics. 

\begin{figure*}
\center
\includegraphics[width=0.89\textwidth,trim=0 20 0 40,clip]{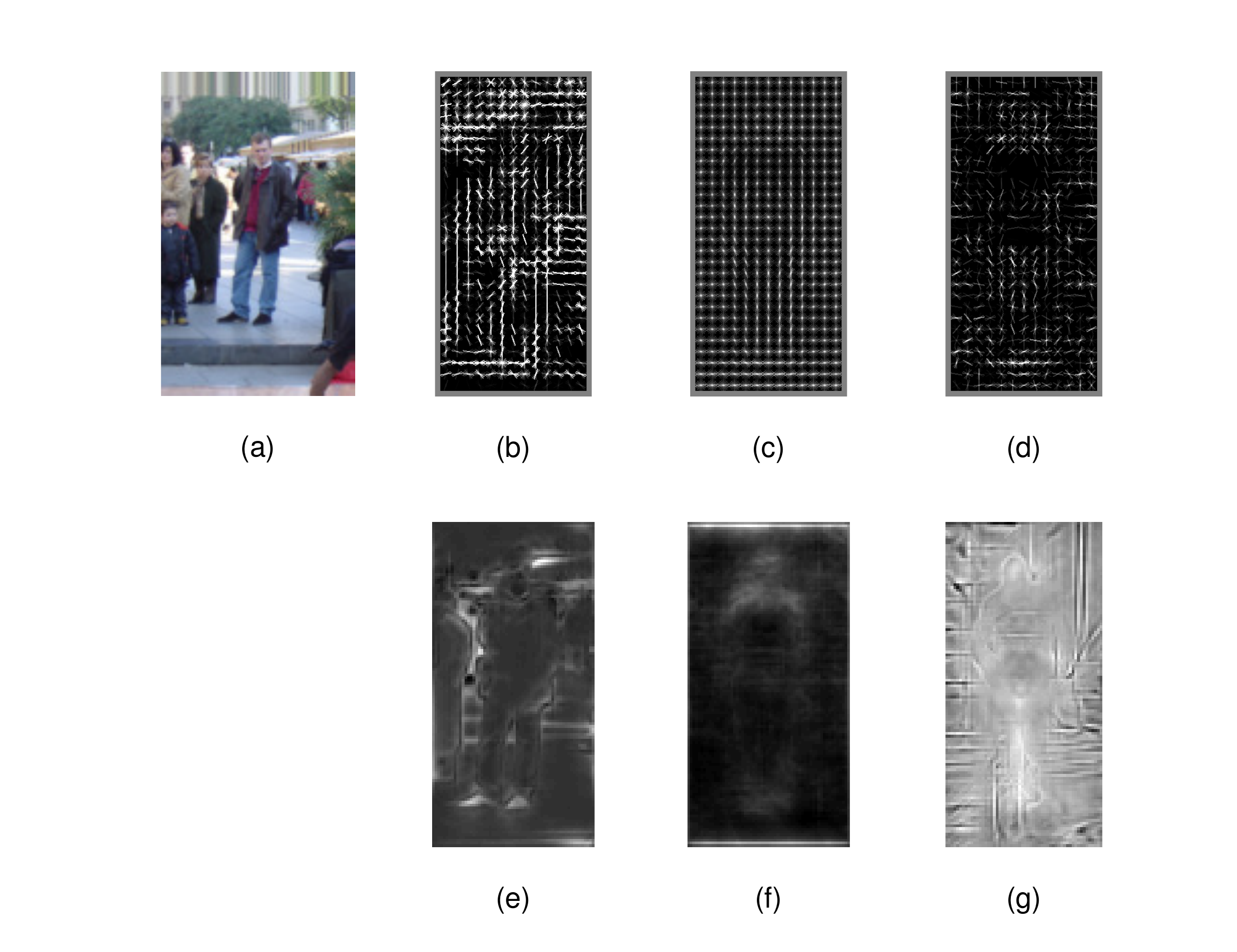}
\caption{Visualisations of HOG and local quadratic classifiers on the INRIA person dataset. (a) A sample image from the set and its equivalent representation in (b) HOG space, and (e) local quadratic space. Lighter shades represent strong edges and correlations respectively. The positive training set mean in (c) HOG space and (f) local quadratic space. Positive weights of the final classifiers in (d) HOG space and (g) local quadratic space. In both HOG and local quadratic space, the visualization of individual images shows significant extraneous informative, however the final classifiers are clearer. Interpreting the local quadratic classifier in (g) is difficult since the correlations cannot be interpreted as edges, however the distribution of positive weights is similar to HOG, especially around the head and shoulders and between the legs.
\label{fig:inria}}
\end{figure*}

\section{Discussion}
We began the piece on the premise that HOG features embody interactions requisite to good recognition performance. Their popularity and success on current object recognition benchmarks supports this, at least empirically. Expanding on the work of~\cite{bristow_ECCV_2012}, we showed that HOG features can be reformulated as an affine weighting on the margin of a quadratic kernel SVM. One can take away two messages from this: (i) a quadratic classifier has sufficient capacity to enable discrimination of visual object classes, and (ii) the actual image prior is captured by the weighting matrix. We performed an experiment where classifiers were tasked with discriminating between ``natural'' and ``noise'' images, and found that the quadratic classifier preserving only local pixel interactions was able to separate the two classes, suggesting that the structure of natural images can be exploited by local second-order statistics.
Armed with only these principles, we set out to discover whether it was possible to learn an accurate classifier in a controlled expression recognition setting, and quantify how much data was required for varying amounts of geometric misalignment. \fig{fig:main} illustrates that with enough synthesized data, a local quadratic classifier can learn non-trivial pixel interactions necessary for predicting expressions. Finally, we applied these insights to a pedestrian detection task, and show in \fig{fig:inria-results} that a significant fraction of HOG's performance can be attributed to preserving local second-order pixels interactions, and not the image specific prior (\ie edges) that it encodes. Inspecting the local quadratic classifier visualization from \fig{fig:inria}, one can see that emphasis (strong positive support weights represented by lighter shades) is placed on the object boundaries - specifically around the head and shoulders and between the legs - just as HOG does.


\section{Conclusions}
Linear SVMs trained on HOG features have pervaded many visual perception tasks. We hypothesized that preserving \emph{local second-order interactions} are at the heart of their success. This is motivated by similar findings within the human visual system. With these simple assumptions combined with large amounts of training data, it is possible to learn a classifier that performs well on a constrained expression recognition task, and within ballpark figures of a HOG-based classifier tailored specifically to an image prior on a pedestrian detection experiment. As the size of datasets continues to grow, we will be able to rely less and less on prior assumptions, and instead let data drive the models we use. Local second-order interactions are one of the simplest encoders of natural image statistics that ensure such models have the \emph{capacity} to make informed predictions.

\bibliographystyle{ieee}
{\small\bibliography{citations}}

\end{document}